\documentclass[final]{cvpr}

\usepackage{times}
\usepackage{epsfig}
\usepackage{graphicx}
\usepackage{amsmath}
\usepackage{amssymb}
\usepackage{subcaption}

\usepackage[pagebackref=true,breaklinks=true,colorlinks,bookmarks=false]{hyperref}

\def\cvprPaperID{****} 
\def\confYear{CVPR 2021}

\begin{document}

\title{EEV: A Large-Scale Dataset for Studying Evoked Expressions from Video}

\author{Jennifer J. Sun$^{1,}$\thanks{} \and Ting Liu$^{2}$ \and Alan S. Cowen$^{2}$ \and Florian Schroff$^{2}$ \and Hartwig Adam$^{2}$ \and Gautam Prasad$^{2}$ \\ \\
\hspace{1cm}$^{1}$\text{Caltech} \hspace{2cm}$^{2}$\text{Google Research}
}

\twocolumn[{%
\renewcommand\twocolumn[1][]{#1}%
\maketitle
\vspace{-0.4in}
\begin{center}
    \centering
    \includegraphics[width=0.88\textwidth]{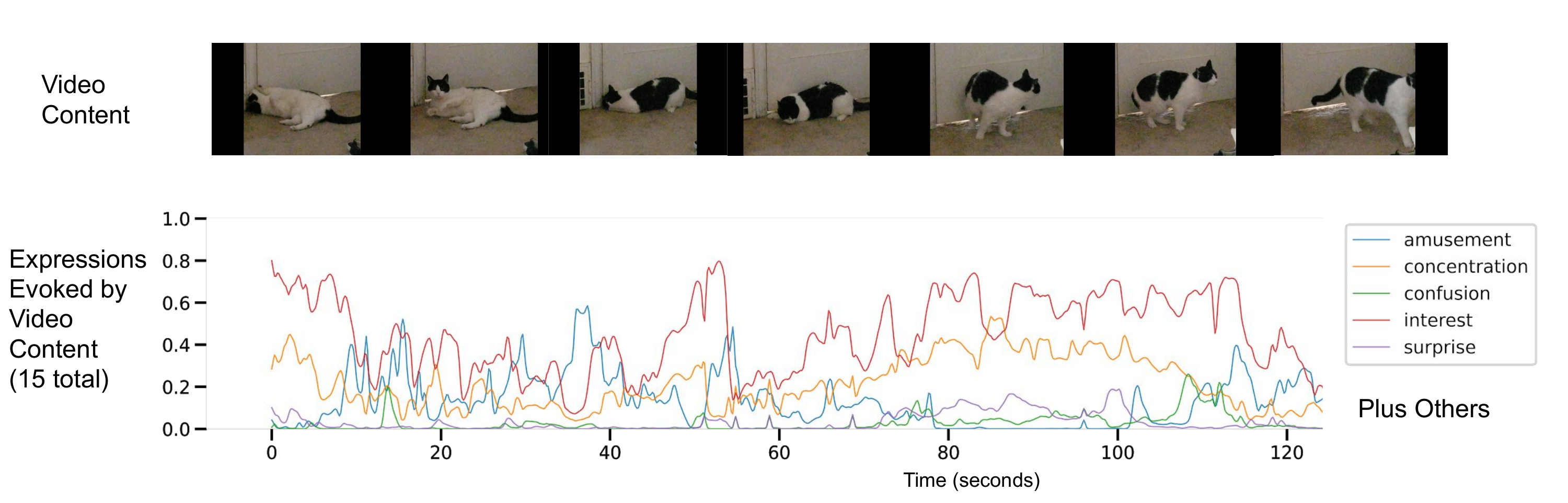}
    \captionof{figure}{The EEV dataset contains publicly available content videos alongside detected viewer facial expressions while viewers are watching the video. There are 15 expression labels annotated at 6Hz over each content video. }
\end{center}%
}]


{
  \renewcommand{\thefootnote}%
    {\fnsymbol{footnote}}
  \footnotetext[1]{This work was done while the author was a research intern at Google.}
}

\begin{abstract}
Videos can evoke a range of affective responses in viewers. The ability to predict evoked affect from a video, before viewers watch the video, can help in content creation and video recommendation. We introduce the Evoked Expressions from Videos (EEV) dataset, a large-scale dataset for studying viewer responses to videos. Each video is annotated at 6 Hz with 15 continuous evoked expression labels, corresponding to the facial expression of viewers who reacted to the video. We use an expression recognition model within our data collection framework to achieve scalability. In total, there are 36.7 million annotations of viewer facial reactions to 23,574 videos (1,700 hours). We use a publicly available video corpus to obtain a diverse set of video content. We establish baseline performance on the EEV dataset using an existing multimodal recurrent model. Transfer learning experiments show an improvement in performance on the LIRIS-ACCEDE video dataset when pre-trained on EEV. We hope that the size and diversity of the EEV dataset will encourage further explorations in video understanding and affective computing. A subset of EEV is released at \href{https://github.com/google-research-datasets/eev}{https://github.com/google-research-datasets/eev}.
\end{abstract}



\section{Introduction}
Videos can be described by their semantic content and affective content. The semantic content focuses on ``What is in the video?'', while the affective content focuses on ``What does the video make people feel?''~\cite{soleymani2014corpus}. Our work focuses on understanding the affective content evoked by visual and audio information from videos. We introduce a scalable framework to annotate viewer expressions evoked by video content, and release the Evoked Expressions from Videos (EEV) dataset to study these evoked expressions.

Recent studies have shown promising results in describing videos using their semantic content, and in parallel, we would like to work on understanding how videos evoke different affect in viewers. For semantic content, video datasets such as Sports1M~\cite{karpathy2014large}, Kinetics~\cite{carreira2017quo}, YouTube8M~\cite{yt8m}, and Moments in Time~\cite{monfort2019moments} have provided benchmarks for performance and enabled advances in semantic video understanding. Currently, datasets for studying viewer affective responses to video content are much smaller~\cite{koelstra2012deap, baveye2015liris, soleymani2012multimodal} and typically focus on one video type, such as films. These limitations can be attributed to the challenge of collecting frame-level affective labels in videos. It is difficult to find reliably affect-evoking stimuli~\cite{horvat2013multimedia} and affective labels are more subjective given they depend on viewer background and context~\cite{soleymani2014corpus}.

\begin{figure}
\centering
\begin{subfigure}{\columnwidth}
    \includegraphics[width=0.8\linewidth]{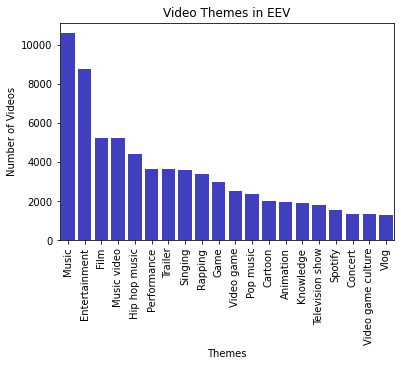}
    \caption{Distribution of the top 20 themes in EEV.}  
    \label{fig:themes_a}
\end{subfigure}

\begin{subfigure}{\columnwidth}
  \includegraphics[width=\linewidth]{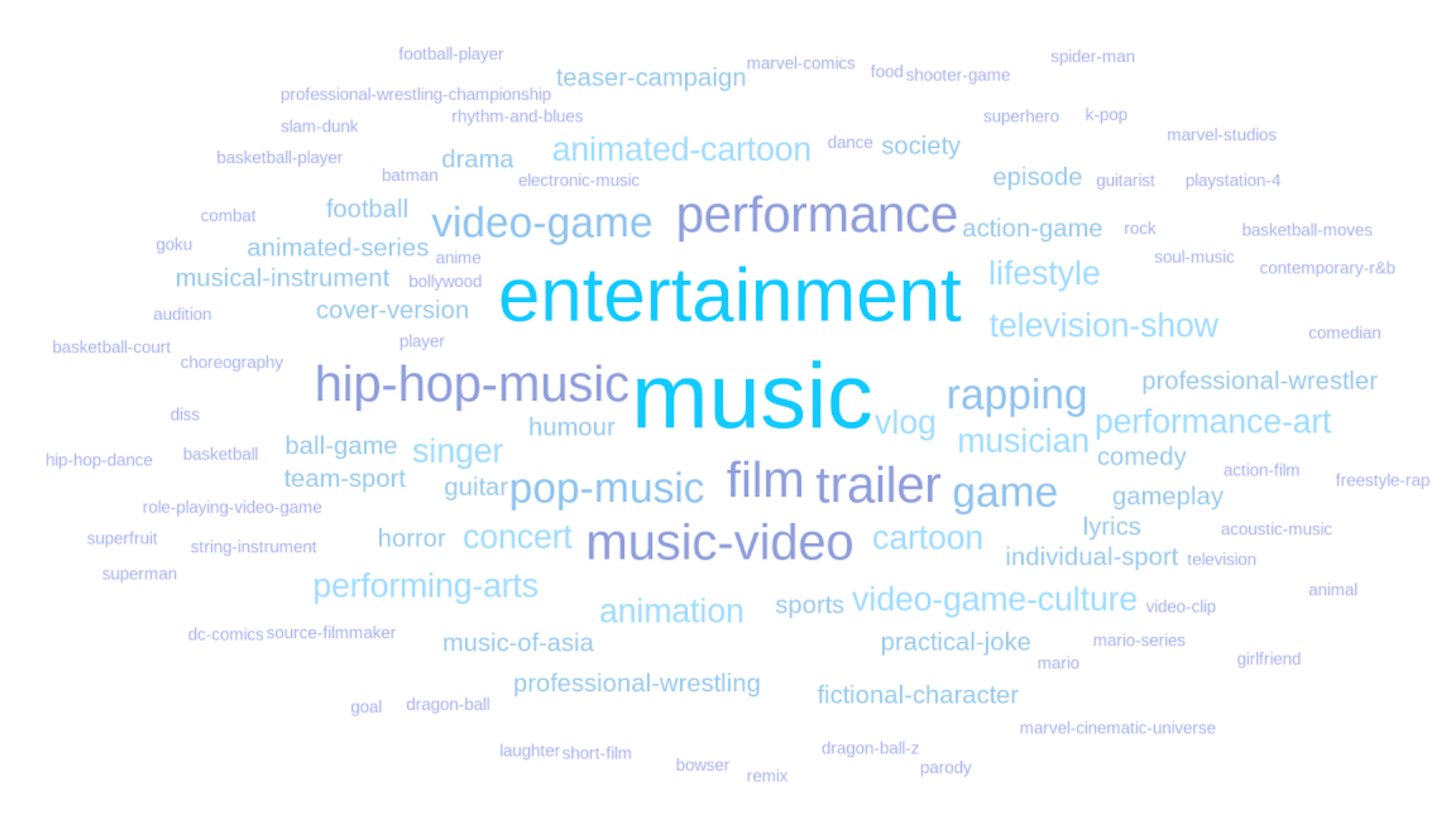}
      \caption{Word cloud of video themes in EEV.}
    \label{fig:themes_b}
\end{subfigure}

  \caption{Video themes in EEV annotated using an automatic system described in section \ref{sec:video_themes}. The themes are not mutually exclusive.}
  \label{fig:themes}
\end{figure}

To build a large-scale dataset for understanding affective content, we introduce a scalable method for annotating viewer expressions evoked by video content by using a video-based model to annotate facial expressions in our data collection pipeline.
The expression recognition model we use is shown to have higher correlation to the population average than single human annotators (Section \ref{sec:expression_annotation}). We use publicly available videos with associated viewer reactions to generate evoked expression labels. The result is the EEV dataset, with 23,547 videos densely annotated at 6 Hz for a total of 36.7 million annotations. The diversity of our dataset is shown in Figure~\ref{fig:themes}. To the best of our knowledge, the EEV dataset is currently the largest dataset for studying affective responses to video from evoked viewer facial expressions. 

The size and diversity of our dataset enables us to study unique questions related to affective content. These include: how well can we predict evoked viewer facial expressions directly from video content?; are some evoked facial expressions easier to predict?; and how well do video themes correspond to different expressions? We explore these areas by analyzing the characteristics of the EEV dataset and establishing baseline expression prediction benchmarks.

Our contributions in this paper are:
\begin{itemize}
    \item The Evoked Expressions from Videos (EEV) dataset, which is a large-scale dataset annotated at 6 Hz for studying evoked expressions from diverse videos. There are 15 annotated expressions. A subset of EEV is publicly available at \href{https://github.com/google-research-datasets/eev}{https://github.com/google-research-datasets/eev}.
    \item A scalable method for annotating evoked viewer facial expressions in online videos.
    \item A baseline model based on~\cite{sungla} and performance benchmark for predicting evoked expressions using the EEV dataset.
\end{itemize}

\section{Related Work}

\subsection{Affective Video Content Analysis} 

Our work is closely related to the field of affective video content analysis, which aims to predict viewer reactions evoked by videos \cite{baveye2018affective,wang2015video}. Hanjalic and Xu \cite{hanjalic2005affective} proposed one of the earliest efforts in the field, mapping movies to continuously evoked valence and arousal. Often, both viewer and video information are incorporated to predict video affect \cite{soleymani2009bayesian, soleymani2012multimodal, koelstra2012deap, becker2017emotion, soleymani2012multimodal}. In these works, physiological signals are measured from a viewer as they watch the video, and combined with visual features for prediction. Other studies \cite{baveye2015liris,jiang2014predicting,zhang2009utilizing,zlatintsi2017cognimuse} use the direct approach: they aim to predict viewer responses directly from video content. Given that we do not use any additional viewer information, our work most closely relates to the direct approach.

Models in this area are often multimodal, given that both visual and auditory features contribute to a video's affective content \cite{wang2006affective}. A review paper \cite{wang2015video} observed that modeling frameworks often consist of video feature extraction, modality fusion, and classification or regression (depending on the model of affect). Recent models generally follow this framework using neural networks \cite{poria2015towards,thugla,sungla,acar2017comprehensive,baveye2015deep,kahou2013combining}. In particular, the LIRIS-ACCEDE dataset \cite{baveye2015deep,baveye2015liris}, part of the MediaEval benchmark \cite{dellandrea2018mediaeval} provides a way to compare model performances for affective content analysis. The top performing models \cite{thugla,sungla} in the MediaEval competition applied RNNs to model the temporal aspects of video with a  multi-modal approach. We establish the performance baseline on the EEV dataset based on these top performing models from the MediaEval benchmark.

\begin{table*}
\small
\begin{center}
\begin{tabular}{l c c c c c c c}
\hline
Dataset & Source & Annot. Type & Annot. Freq & Num Videos \\
\hline
COGNIMUSE \cite{zlatintsi2017cognimuse} & movies & affective labels (valence, arousal, etc.) & frame & 7   \\
HUMAINE \cite{douglas2007humaine} & selected clips &  affective labels (valence, arousal, etc.) & global and frame & 50  \\
FilmStim \cite{schaefer2010assessing} & movies & affective labels (valence, arousal, etc.) & global and frame & 70 \\
DEAP \cite{koelstra2012deap} & music videos & affective labels (valence, arousal, etc.), face video & frame & 120 \\
VideoEmotion \cite{jiang2014predicting} & online videos & discrete emotions & global & 1101 \\
LIRIS-ACCEDE (discrete) \cite{baveye2015deep,baveye2015liris} & movies & valence, arousal & frame & 160 \\
LIRIS-ACCEDE (MediaEval) \cite{baveye2015deep,baveye2015liris} & movies & valence, arousal & frame & 66  \\
\hline
EEV (ours) & online videos & evoked expressions & frame & 23,574 \\
\hline
\end{tabular}
\end{center}
\caption{Dataset size comparison for predicting viewer affect from video content.}
\label{table:dataset}
\end{table*}

\paragraph{Facial response to media}. Videos have often been used to evoke viewer facial expressions in studies across psychology and affective computing \cite{ck2017,baveye2018affective, horvat2013multimedia,mcduff2013affectiva,soleymani2016analysis} (e.g.\ choosing a funny video so that viewers laugh). Our corpus of videos can be used to identify content that evoke distinct facial expressions and thus may be useful for these studies. 

Facial expressions have been also used as predictors in studies on self-reported emotions~\cite{soleymani2016analysis,mcduff2017large}, facial landmark locations~\cite{deng2017factorized} and viewer video preferences \cite{zhao2013video,mcduff2015predicting,mcduff2013predicting}. These studies have also used automated systems to annotate facial expressions in videos. However, rather than using facial expressions as the predictors, we predict evoked viewer facial expressions directly from online videos. 

\paragraph{Facial expression recognition.} The expressions in the EEV dataset are annotated using a facial expression recognition model similar to \cite{Vemulapalli_2019_CVPR}, with expressions based on \cite{ck2019}. Traditionally, automated methods predict facial action units from the facial action coding system (FACS) \cite{ekman1978manual,martinez2017automatic}. However, it remains difficult to obtain accurate FACS annotations in natural contexts, such as user uploaded videos, that have uncontrolled conditions of viewpoint, lighting, occlusion, and demographics. We instead apply a semantic space approach to classifying facial expressions, which was recently introduced by \cite{ck2019}. This approach has revealed that in natural images, people reliably recognize a much wider array of facial expressions than have traditionally been studied using FACS. Information conveyed by facial expressions can be represented efficiently in terms of categories such as ``amusement" and ``surprise".

\subsection{Datasets for Predicting Affective Responses}
Datasets with viewer affective response annotations are important for affective video content analysis. The size of current datasets for predicting viewer affective response to video are small relative to the size of other video benchmark datasets \cite{yt8m,karpathy2014large,carreira2017quo}. Existing datasets such as DEAP \cite{koelstra2012deap}, VideoEmotion \cite{jiang2014predicting}, and FilmStim \cite{schaefer2010assessing} are labelled with one annotation per video clip, and the largest (VideoEmotion) contains 1,101 videos. While these datasets are useful for other applications, they are too small to test complex models, and cannot be used to understand temporal changes in affect. Some datasets annotate evoked viewer affect over time, such as LIRIS-ACCEDE \cite{baveye2015deep,baveye2015liris}, COGNIMUSE \cite{zlatintsi2017cognimuse}, and HUMAINE \cite{douglas2007humaine}. These have frame level annotations based on viewer self-reports. The largest, LIRIS-ACCEDE \cite{baveye2015deep,baveye2015liris}, consists of 160 films with annotations every second. A subset of LIRIS-ACCEDE (66 films) is annotated continuously with valence and arousal from self-report used in the MediaEval benchmark \cite{dellandrea2018mediaeval}. These datasets often focus on one category of video (films or music videos).

Compared to existing datasets, EEV is significantly larger, as shown in Table~\ref{table:dataset}. To achieve scale, we use a facial expression recognition model.
Our evaluations in Section~\ref{sec:expression_annotation} shows that this model has higher correlations to the population mean than human raters on average. 
Although we use facial expressions instead of self-report, both methods for measuring affect have been found to share significant variance \cite{KassamKarim2011Aoee}. The results by \cite{KassamKarim2011Aoee} also found that analyzing facial expressions can provide unique insights into affective experiences. Another characteristic of the EEV dataset is that it contains diverse video themes, shown in Figure ~\ref{fig:themes}, enabling affective content to be studied across categories.

\paragraph{Subjectivity.} We recognize that there is a need for personalization in understanding affective content due to the fact that viewer response depends on the experience of the subject \cite{baveye2018affective,wang2015video,soleymani2014corpus}. However, as we observe and as argued by \cite{soleymani2014corpus}, affective response is not arbitrary, and often agreement can be found across viewers. The challenge, as identified by \cite{soleymani2014corpus}, is to recognize common affective triggers in videos. By providing the largest video dataset annotated with evoked viewer facial expressions, EEV enables this challenge to be explored further in the future.


\section{Dataset Analysis} \label{sec:dataset_char}

\begin{figure}
\centering
  \includegraphics[width=\linewidth]{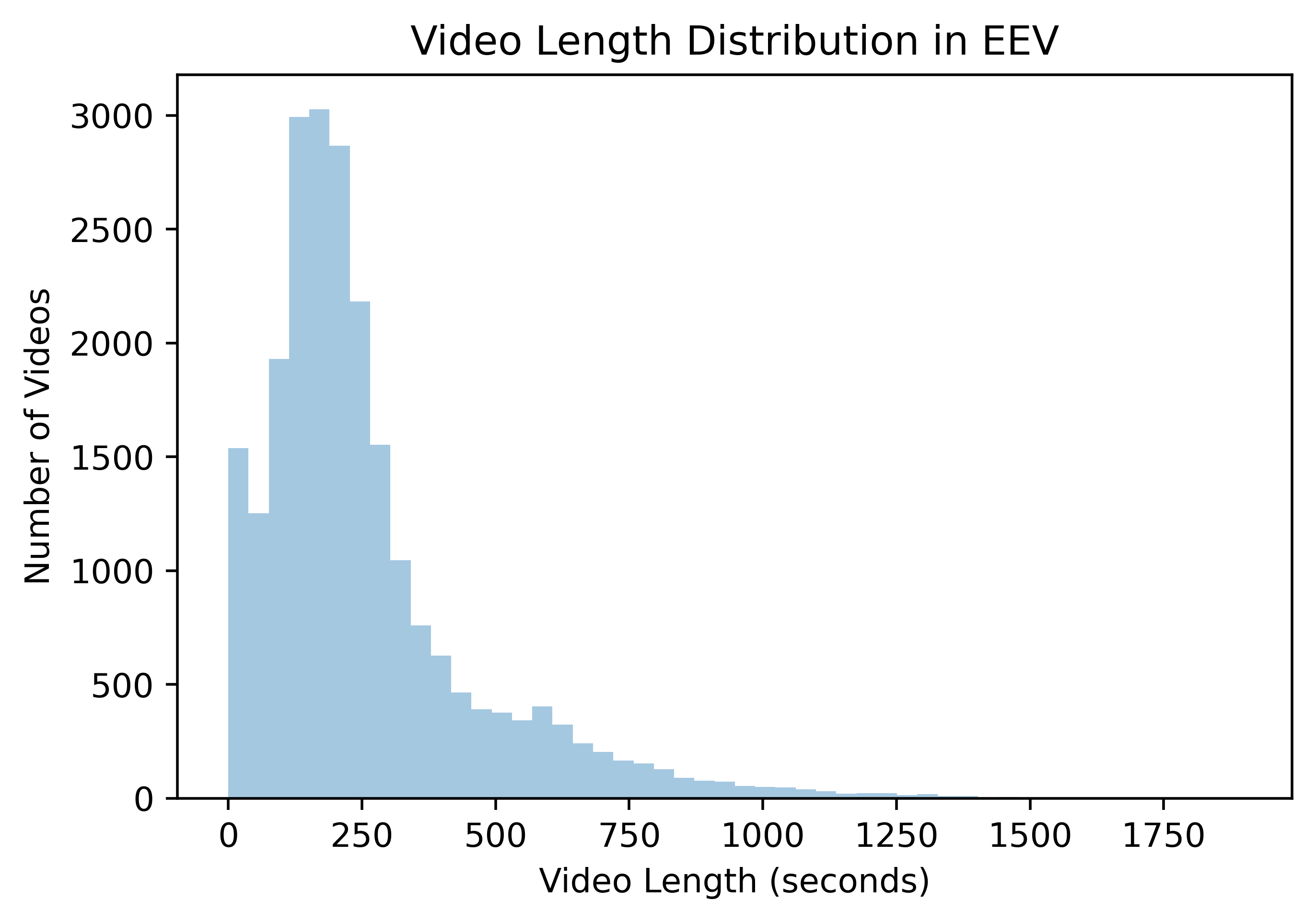}
  \caption{The distribution of video lengths in the EEV dataset. The average video length is 4.3 minutes.}
  \label{fig:lengths}
\end{figure}

The EEV dataset consists of 23,574 videos annotated at 6 Hz for a total of 36.7 million automatic annotations. The video length distribution is show in Figure~\ref{fig:lengths} and there are a total of 1,700 hours. The annotations are over 15 expression classes from \cite{ck2019}: amusement, anger, awe, concentration, confusion, contempt, contentment, disappointment, doubt, elation, interest, pain, sadness, surprise, and triumph. Each annotation is a 15-dimensional vector with values ranging from 0 to 1, corresponding to the confidence of each expression. 

\begin{figure}
\centering
  \includegraphics[width=\linewidth]{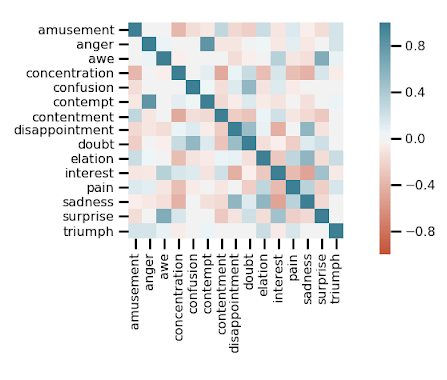}
  \caption{The correlation matrix between the evoked expressions in EEV (Best viewed in color).}
  \label{fig:correlation}
\end{figure}

\subsection{Expressions in EEV}
The expressions in EEV are based on \cite{ck2017, ck2019}. The study on naturalistic facial expressions, \cite{ck2019}, has a more detailed analysis of each expression. In particular, Table~\ref{table:exp}, reproduced from with permission \cite{ck2019}, shows the free-response terms from a user study most correlated with each expression. 

\begin{table}
\begin{center}
\small
\begin{tabular}{l c}
\hline
Expression & Correlated Terms \\
\hline
Amusement & happiness, laughter, extreme happiness \\
Anger & boiling with anger, angry contempt, feeling mad \\
Awe & surprise, awestruck surprise, wonder \\
Concentration & deep focus, determination, focus  \\
Confusion & feeling perplexed, bewilderment, dumbfoundedness \\
Contempt & annoyance, disapproval, distrust  \\
Contentment & relaxation, peacefulness, calmness \\
Disappointment & sadness, regret, frustration \\
Doubt & distrust, suspicion, contemptous doubt \\
Elation & extreme happiness, happiness, excitement, laughter \\
Interest & childlike curiosity, curiosity, wonder \\
Pain & severe pain, angry pain, feeling hurt \\
Sadness & extreme sadness, crying, feeling upset \\
Surprise & shock, awestruck surprise, extreme surprise \\ 
Triumph & excitement, great triumph, surprise \\
\hline
\end{tabular}
  \caption{Free-response terms correlated with each expression category from \cite{ck2019}, which are also categories of evoked expressions studied in EEV.}\label{table:exp}
\end{center} 
\end{table}

\subsection{Evoked Expression Labels}
\paragraph{Expression distribution.} The distribution of each expression over videos is plotted in Figure~\ref{fig:full_distribution}. We observe that the distribution varies across expressions. Some expressions, such as anger and contempt, have relatively rare high confidence samples, while other expressions, such as concentration and interest, have relatively common high confidence samples. This property makes EEV an interesting dataset for studying regression on labels of different distributions, or classification (by binarizing expressions using a threshold).


\def\figsize{0.2}
\def\fighspace{-3mm}
\def\fighspacer{-3mm}
\begin{figure*}
\centering
\begin{tabular}{ccccc}
\centering

\includegraphics[width=\figsize\textwidth]{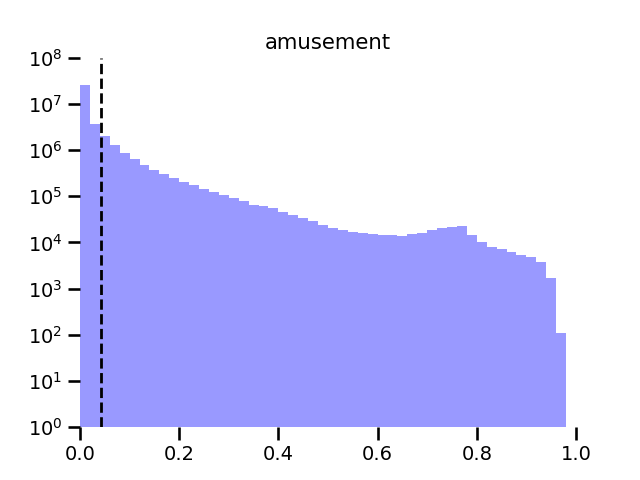}\hspace{\fighspace} & 
\includegraphics[width=\figsize\textwidth]{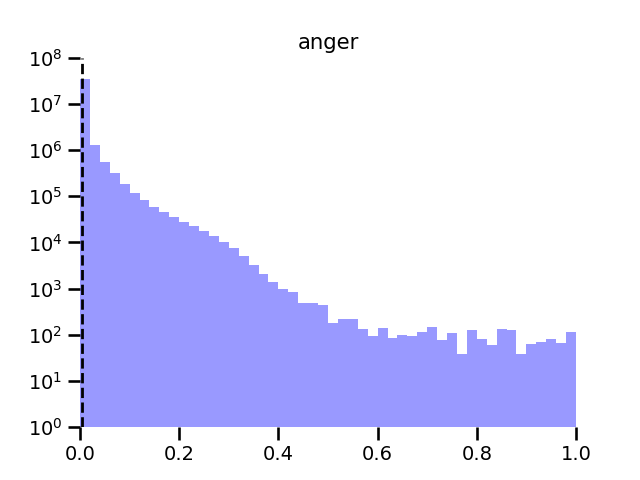}\hspace{\fighspacer} & 
\includegraphics[width=\figsize\textwidth]{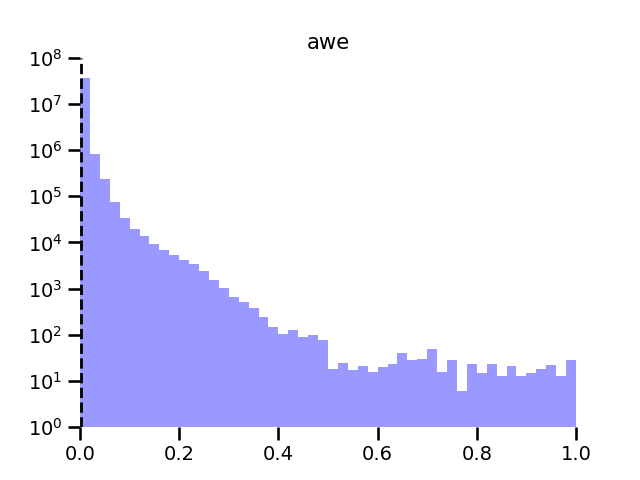}\hspace{\fighspace} & 
\includegraphics[width=\figsize\textwidth]{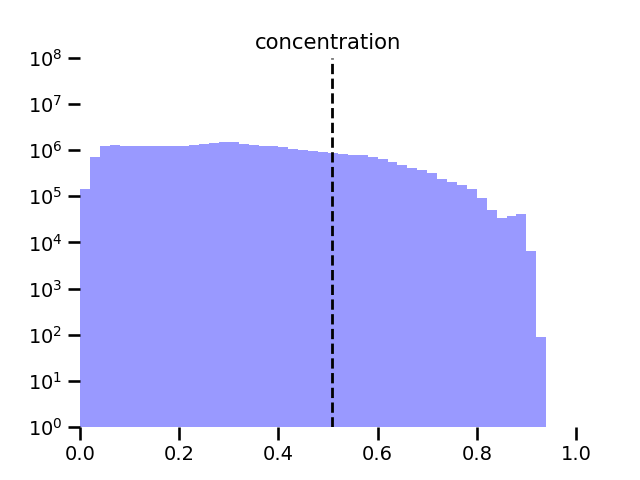}\hspace{\fighspacer} & \includegraphics[width=\figsize\textwidth]{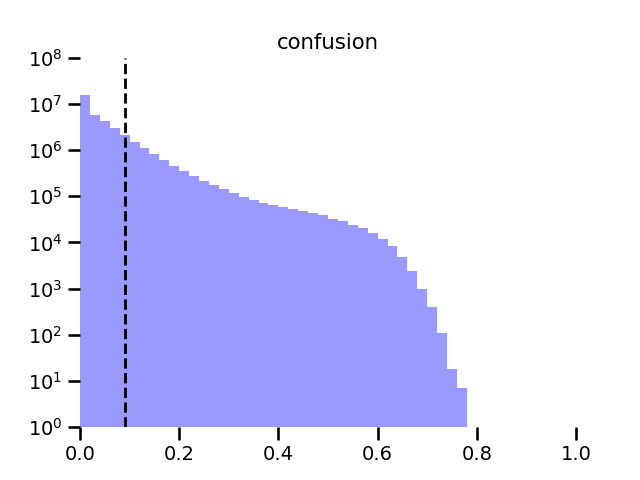}\hspace{\fighspace} \\

\includegraphics[width=\figsize\textwidth]{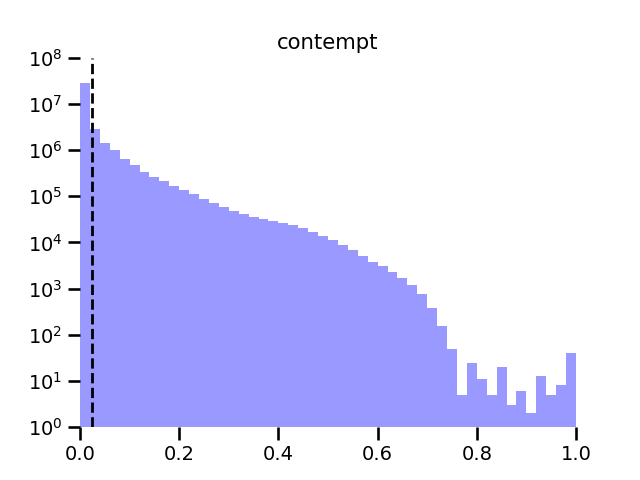}\hspace{\fighspacer}  & \includegraphics[width=\figsize\textwidth]{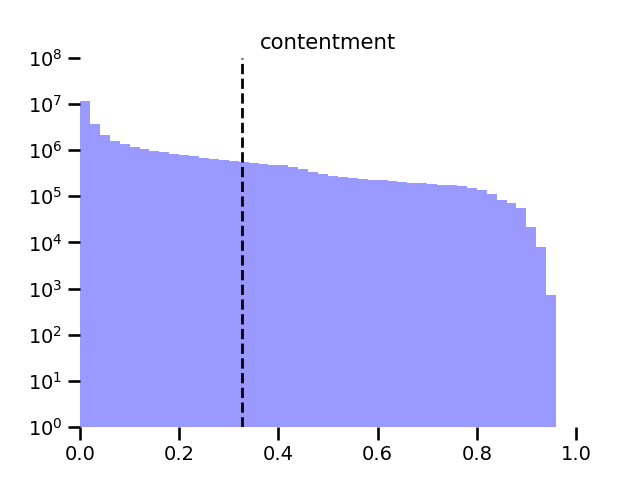}\hspace{\fighspace} & \includegraphics[width=\figsize\textwidth]{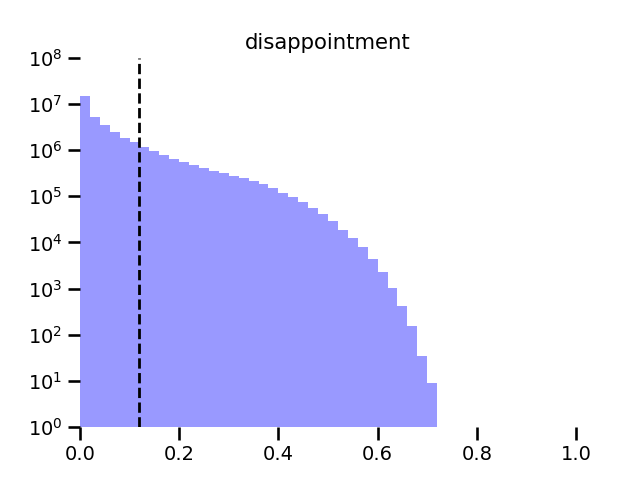}\hspace{\fighspacer} &  \includegraphics[width=\figsize\textwidth]{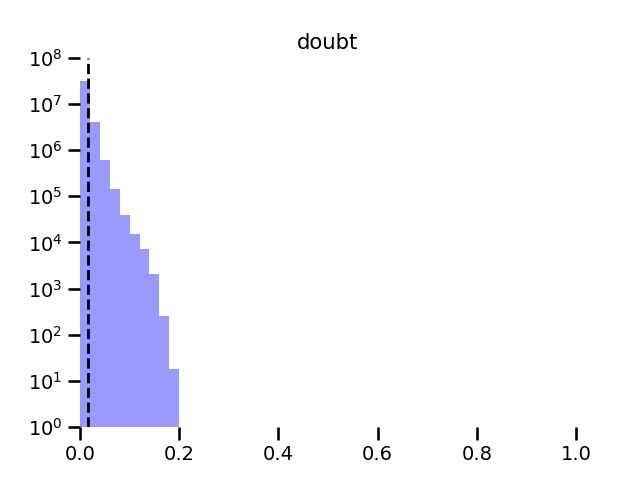}\hspace{\fighspace} & \includegraphics[width=\figsize\textwidth]{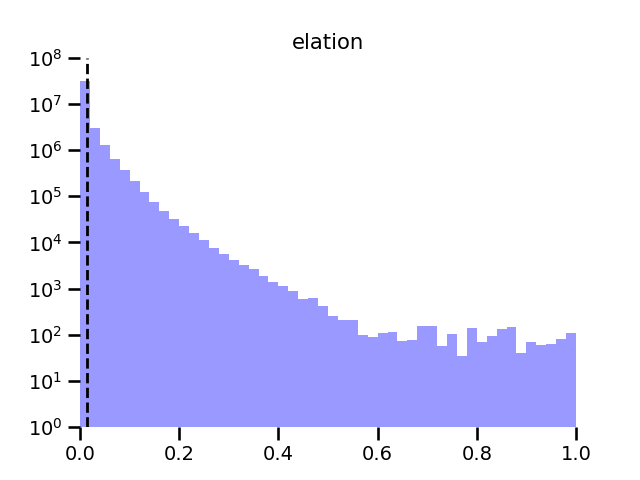}\hspace{\fighspacer}  \\

\includegraphics[width=\figsize\textwidth]{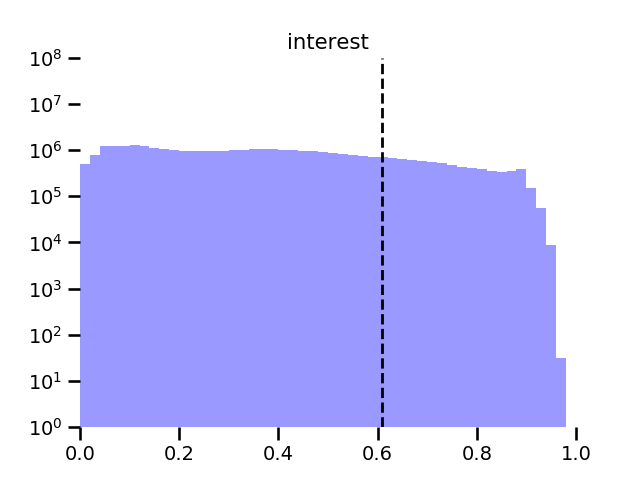}\hspace{\fighspace} & \includegraphics[width=\figsize\textwidth]{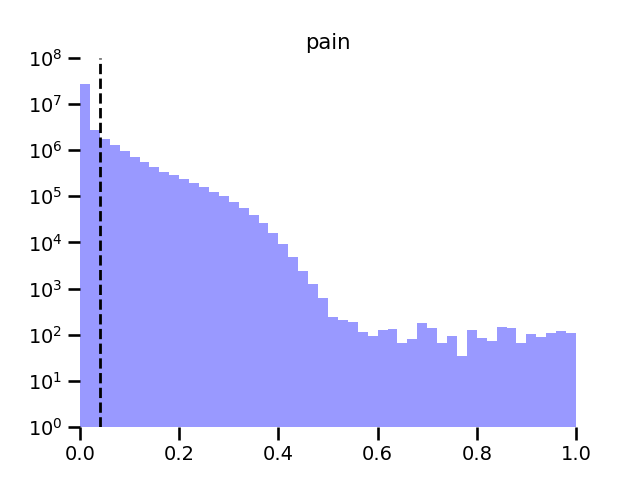}\hspace{\fighspacer} & s\includegraphics[width=\figsize\textwidth]{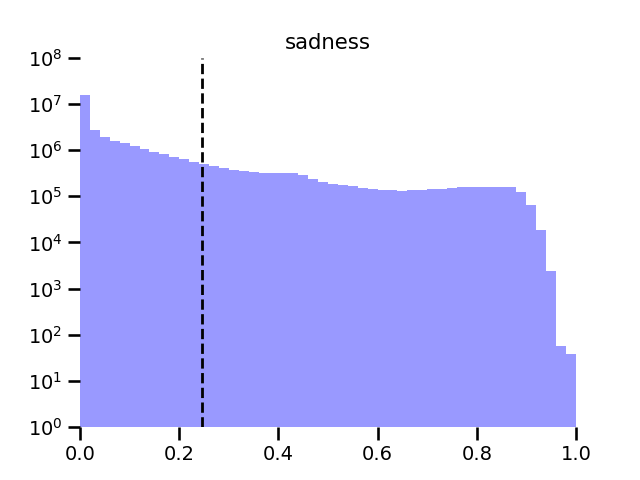}\hspace{\fighspace} & \includegraphics[width=\figsize\textwidth]{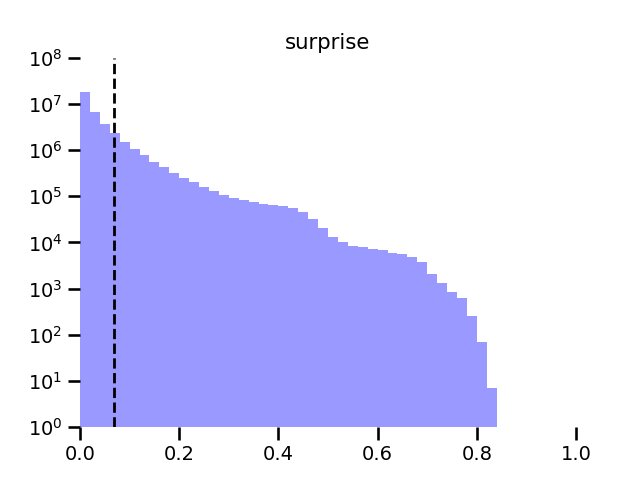}\hspace{\fighspacer}  & \includegraphics[width=\figsize\textwidth]{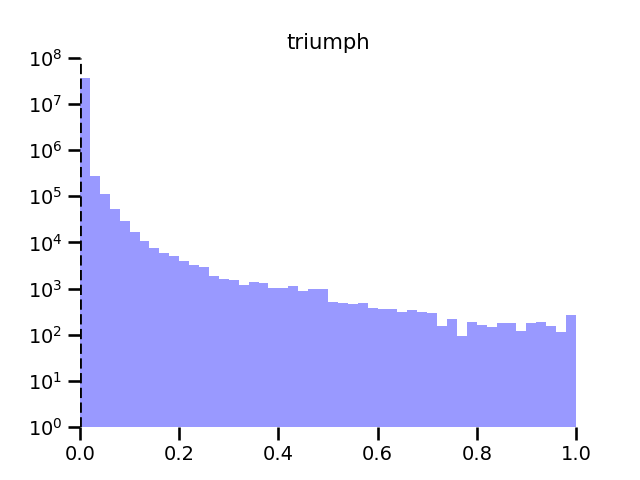}\hspace{\fighspace} \\

\end{tabular}
\caption{Distribution of all expressions in EEV. The dotted line represents the 80 percentile of the distribution. } \label{fig:full_distribution}
\end{figure*}

\paragraph{Expression correlation.}  Using EEV, we can examine the correlation matrix of different evoked expressions in Figure~\ref{fig:correlation}. There is a positive correlation between expressions associated with happiness, such as amusement, contentment and elation. A strong correlation exists between anger and contempt (associated with disapproval), as well as surprise and awe. For negative correlation, sadness is anti-correlated with interest and concentration. Contentment tends to be negatively correlated with concentration and interest is negatively correlated with disappointment. This correlation is reasonable given the association of each expression with different keywords provided in supplementary materials.

\paragraph{Dataset split.} We split the EEV dataset at the video level with a roughly 60:15:25 split into 18,124 training videos, 4,530 validation videos and 7,551 test videos for machine learning applications. We verified that there are similar distributions of expressions in each dataset split.

\paragraph{Challenges.} The EEV dataset can be challenging because the viewer expression can depend on viewer background, external context and other information not present in the video from visual and audio based data. This is a challenge for directly predicting viewer response from general, ``in-the-wild" videos. Despite this, we show that our baseline model can learn useful information from the EEV dataset using only video content in section~\ref{sec:eev_results}.

\subsection{Video Themes} \label{sec:video_themes}

\paragraph{Theme annotation.} We characterize the EEV dataset in terms of video themes (key topics that can be used to describe the video). The video themes are obtained from the video annotation system described in \cite{googleio}. These annotations correspond to the Knowledge Graph entities \cite{kg} of the video, which are computed based on video content and metadata \cite{googleio}. We summarize each video into a set of video themes using the Knowledge Graph entities, similar to the approach used by YouTube8M \cite{yt8m}. This is so that we can better understand the video composition of the EEV dataset. The distribution of the themes in EEV is shown in Figure~\ref{fig:themes}.

\begin{table*}
\begin{center}
\small
\begin{tabular}{l c c }
\hline
Expression & Top 3 Themes & Bottom 3 Themes \\
\hline
Amusement & Comedy, Humour, VG & Horror, Rapping, MV \\
Anger & Trailer, Sports, Gameplay & Singer, Pop Music, Concert  \\
Awe & Rapping, Lifestyle, Game & Concert, Singer, Action  \\
Concentration & Horror, Trailer, Action & Vlog, Humour, Singer \\
Confusion & HH, PA, Action Game & Comedy, Singer, Humor \\
Contempt & Horror, Trailer, Action & Singer, Vlog, Pop music \\
Contentment & Concert, Singer, Comedy & Game, Rapping, Horror  \\
Triumph & Gameplay, Sports, Action game & Singer, Pop, Concert \\
Disappoint. & Knowledge, Concert, PA & Action, Humour, Comedy\\
 Doubt & Horror, Game, Action & Humour, Singer, Comedy \\
Elation & Practical Joke, Vlog, Humour & Trailer, Action, Horror \\
 Interest & PA, Practical Joke, Game & Knowledge, Vlog, Concert \\
Pain & Gameplay, VG, Humour & Lifestyle, PA, Horror \\
Sadness & Practical Joke, Vlog, Singer & Trailer, Horror, Action \\
Surprise & Horror, Lifestyle, MV & Comedy, Singer, Concert\\
\hline
\end{tabular}
  \caption{Distribution of evoked expressions across different video themes in EEV. VG = Video Game, MV = Music Video, PM = Pop Music, PA = Performance Art, HH = Hip Hop}
  \label{table:theme_exp}
\end{center}
\end{table*}

\paragraph{Evoked Expressions and Themes.} We sort each video theme with more than 50 videos by the evoked expressions, and obtain themes that are most and least associated with each expression. This result is shown in Table~\ref{table:theme_exp}. Notably, amusement occurs more in comedy as compared to horror or rapping. Sorting by anger or triumph results in sports and gameplay as two of the top three themes. There tends to be more evoked surprise in horror and lifestyle videos than comedy or singer videos. This association of themes and evoked viewer expressions can be useful for content creators to better understand the affective impact of their videos. Additionally, sorting videos by evoked expressions using EEV can be helpful in selecting videos to evoke specific affective signals for studies.

We hope these unique characteristics of the EEV dataset can encourage further studies in video understanding and affective computing.


\section{Data Collection}

\begin{figure*}
\centering
  \includegraphics[width=0.8\linewidth]{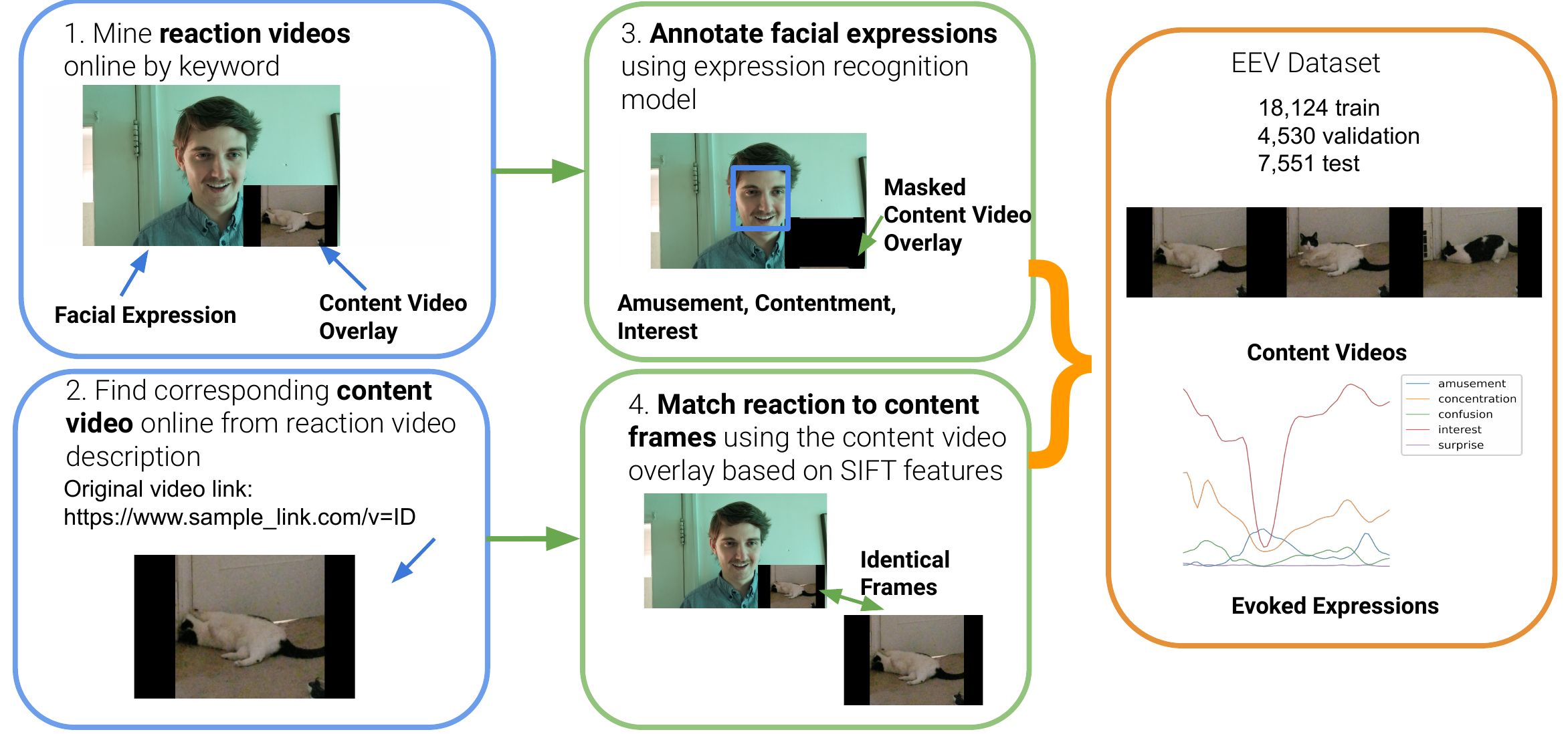}
  \caption{An overview of the data collection process for EEV. }
  \label{fig:3}
\end{figure*}

The EEV dataset leverages facial expression recognition models in order to study affective response to videos at a large scale. The source of the EEV dataset is publicly available reaction videos, which depict viewers reacting to another video. The video that viewers are watching is typically another publicly available video, which we will call the content video. The facial expressions in the reaction video are used to generate evoked expression labels for the content video. An outline is illustrated in Figure \ref{fig:3}. 

\begin{figure*}
\centering

\begin{subfigure}[t]{1.0\columnwidth}
  \includegraphics[width=\linewidth]{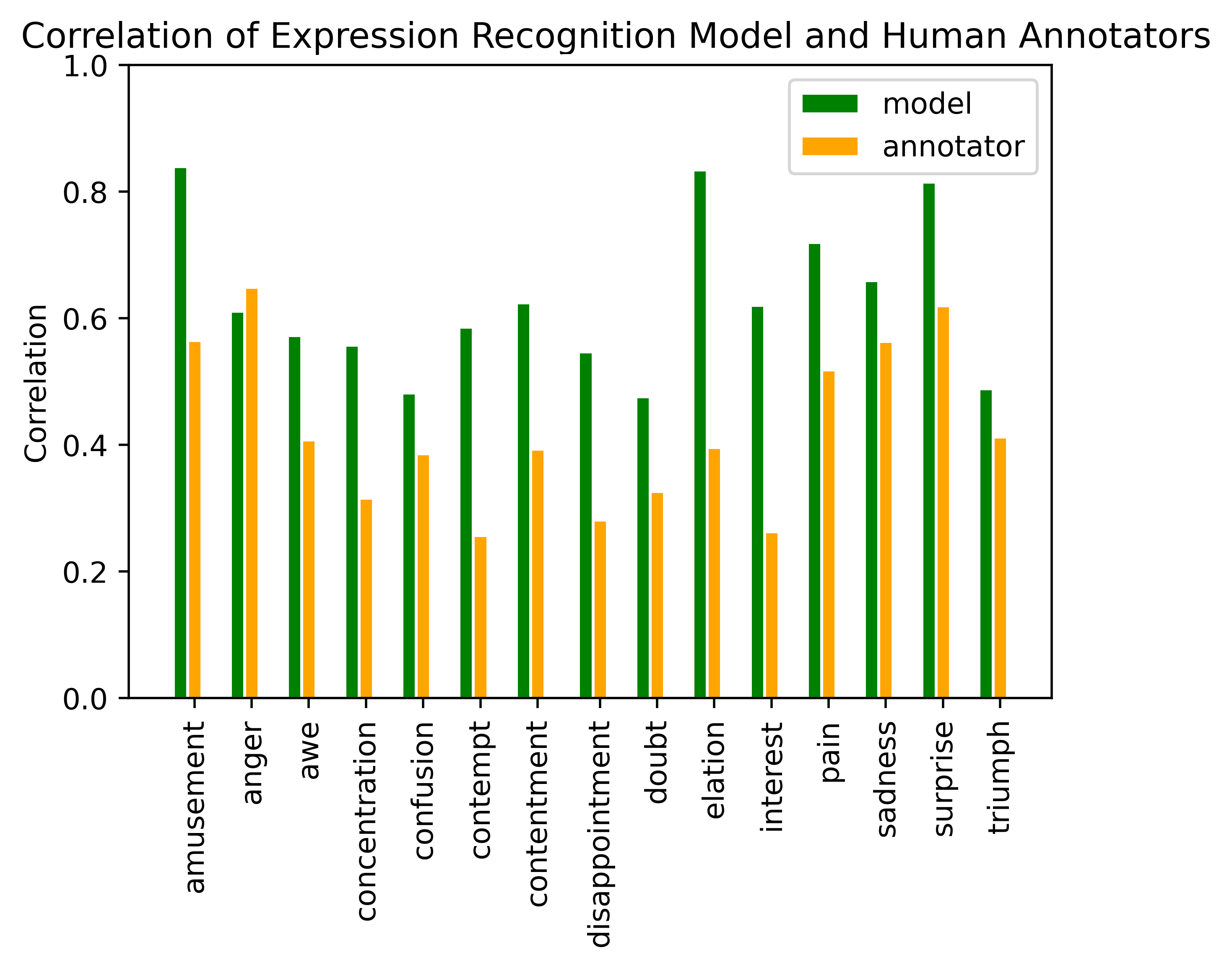}
    \caption{Comparing model predictions to human annotators.}  
    \label{fig:eval_a}
\end{subfigure}
~
\begin{subfigure}[t]{1.0\columnwidth}
  \includegraphics[width=\linewidth]{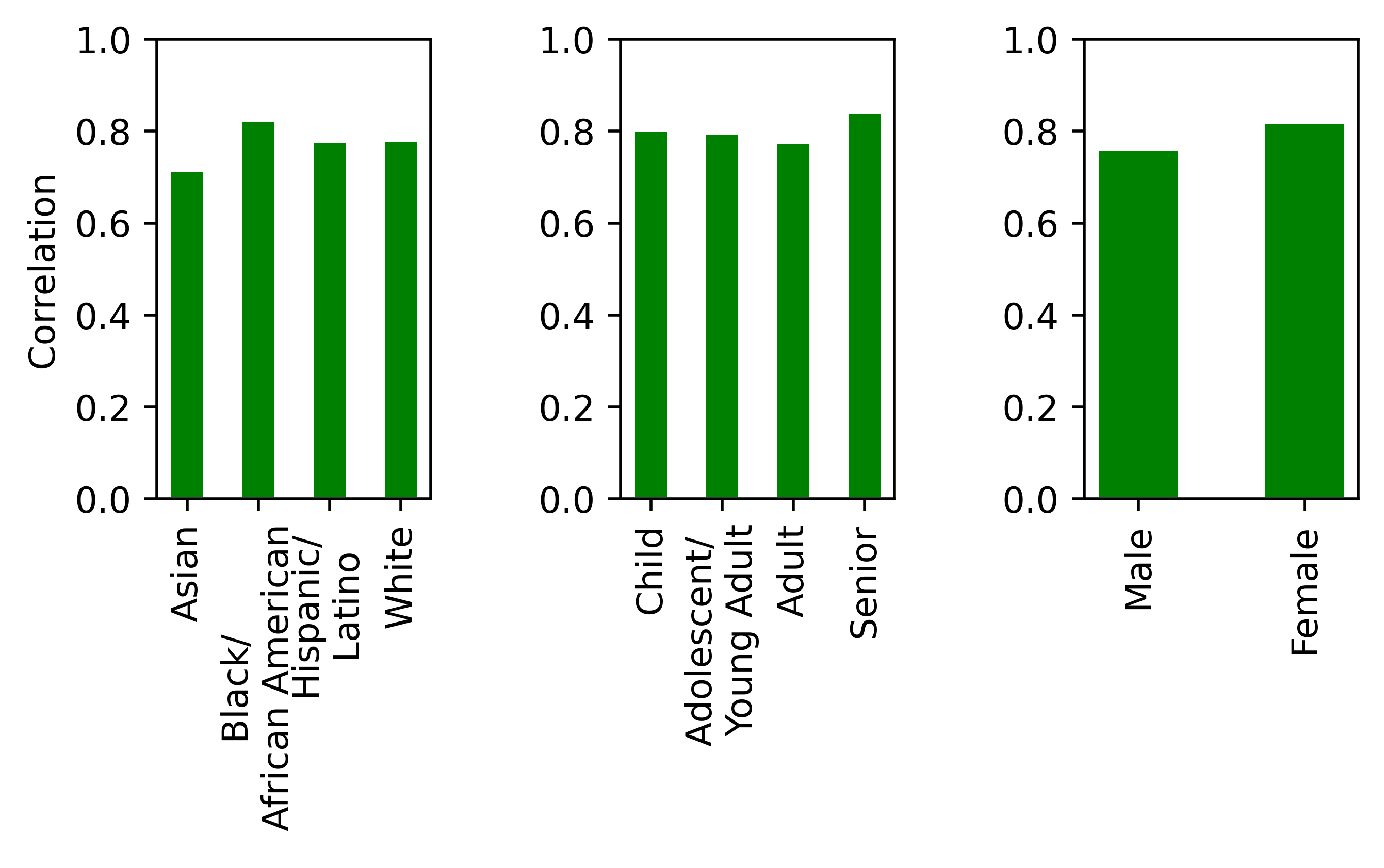}
      \caption{Correlation over different demographics.}
    \label{fig:eval_b}
\end{subfigure}

\caption{Total prediction correlation of the facial expression recognition model on the dataset in \cite{ck2019}, across the sixteen expression classes. }
\label{fig:eval_model}
\end{figure*}

To start, we first compile a list of public reaction videos on an online video corpus by performing a crawl based on keywords in the video title (step 1 in Figure \ref{fig:3}). The reaction videos contain the facial expression of the viewer overlaid on top of the video content they are watching. Using these reaction videos, we find the corresponding content video from the reaction video description (step 2 in Figure \ref{fig:3}). The facial expressions of the viewer in the reaction video is annotated using the model described in Section~\ref{sec:expression_annotation} (step 3 in Figure \ref{fig:3}). Then, the evoked expressions from the reaction video is paired with the content video at each timestamp using SIFT features \cite{lowe2004distinctive} as the facial reactions and content video frame is synchronized (step 4 in Figure \ref{fig:3}). The final dataset consists of the content videos with their evoked expressions from viewer reactions.

\subsection{Facial Expressions} 
Our facial expression categories are based on the work of \cite{ck2017, ck2019}. By analyzing the self-report of viewers from 2,185 emotionally evocative videos, \cite{ck2017} suggests that reported user experiences are better captured by emotion categories rather than dimensional labels. This work defines 27 emotion categories, which they found to be linked by smooth gradients. A followup work \cite{ck2019} examined 28 expression categories that are expressed by face and body using human annotations. The facial expression categories in EEV are based on this work. In our approach, we focus on 16 of the 28 expressions in \cite{ck2019} for which our automatic annotation model performs well. A detailed description of each expression is in supplementary materials and \cite{ck2019}.

\subsection{Facial Expression Annotation}\label{sec:expression_annotation}

EEV is annotated using a facial expression recognition model. We evaluate the automatic method against human annotator agreement on the Berkeley faces dataset \cite{ck2019}. 

\paragraph{Annotation Model.} The facial expression recognition model, similar to \cite{Vemulapalli_2019_CVPR}, uses facial movements over 3 seconds at 6 Hz to predict facial expression labels at each video frame. If more than one face is visible for a frame, we average the model confidences.

In this step, we first compute face-based features using the NN2 FaceNet architecture \cite{Schroff_2015_CVPR} from input video frames. Specifically, we extract the inception (5a) block with a 7x7 feature map composed of 1,024 channels fed into a 7x7 average pooling layer to create a 1,024 dimensional feature vector representing a single face at a single time point in a video. We feed the face features computed over a given video segment into two long short-term memory (LSTM) layers each with 64 recurrent cells to capture temporal information. The output of the LSTM is then fed through a mixture of experts model (2 mixtures in addition to the dummy expert). The network is trained on 274K ratings of 187K clips of faces from a public video corpus. The data is manually annotated by human raters who selected all facial expression categories that applied to each face.

\paragraph{Comparing to Human Annotators.} To evaluate the accuracy of the annotation model on natural expressions, we compare them to human judgments of faces in a held-out test set from the dataset introduced by \cite{ck2019}. We use Pearson's Correlation Coefficient, which varies from -1 (anti-correlated) to 1 (correlated). The results are in Figure~\ref{fig:eval_a} for each expression. We see that the accuracy levels of our model annotations exceed those of single rater human annotations on natural expressions compared with the average human rating in general. Correlations are shown across different demographics groups in Figure~\ref{fig:eval_b}. Importantly, the prediction correlations for our model were similar for faces of different ethnicities, genders, and age groups.


\section{Baseline Model}
Our evoked expression prediction model is based on \cite{sungla}. Our goal is to produce performance baselines for EEV from on an existing model benchmarked on the LIRIS-ACCEDE test set. The LIRIS-ACCEDE dataset \cite{baveye2015deep,baveye2015liris}, part of the MediaEval benchmark \cite{dellandrea2018mediaeval}, provides a way to compare model performances for affective content analysis. The model that achieved top performance on the MediaEval benchmark task \cite{sungla} uses a combination of pre-computed features across multiple modalities, gated recurrent units (GRUs)  \cite{cho2014learning}, and a mixture of experts (MOE). We use a similar architecture in our experiments. This architecture has also been studied by \cite{miech2017learnable} and performed well on the YouTube8M dataset \cite{yt8m}. Figure~\ref{fig:8} presents an overview.

\subsection{Feature Extraction}

Videos present information to viewers through multiple modalities and a multi-modal approach will be needed to understand viewer response \cite{wang2006affective}. We leverage information from image, face, and audio modalities by extracting frame level features for each second in the videos.

\paragraph{Image features.} For the image features, we read the video frames at 1 Hz. We then feed the frames into the Inception V3 architecture \cite{szegedy2016rethinking} trained on ImageNet \cite{deng2009imagenet}. We extract the ReLu activation in the last hidden layer, resulting in a 2048-D feature vector for the image input. 

\paragraph{Face features.} For the face features, we use the same extracted image, but focus on the two largest faces in the image. The face features are extracted using an Inception-based model trained on faces \cite{Schroff_2015_CVPR}. This process results in another 2,048-D feature vector based on the face. We pad the face feature vector with zeros if less than two faces are detected in the image. 

\paragraph{Audio features.} The audio feature extraction is based on the VGG-style model provided by AudioSet \cite{gemmeke2017audio} trained on a preliminary version of YouTube8M \cite{yt8m}. In particular, we extract audio at 16kHz mono and followed the method from AudioSet to compute the log mel-spectrogram. The $96 \times 64$ log mel spectrogram is then fed into the VGG-style model, which outputs 128-D embeddings. 

\subsection{Model Architecture}
\paragraph{Temporal model.} Each feature extracted above (image, face, audio) is fed into its own subnetwork consisting of GRUs. We use GRUs in order to take into account the temporal characteristics of video. The features are extracted at 1 Hz and fed into their respective GRU with a sequence length of 60. The outputs of the final state from each GRU are then concatenated and the fused vector is fed into the regression model. 

\paragraph{Regression model.} We use context gating \cite{miech2017learnable} to weigh the input features before feeding the representation into an MOE model \cite{jordan1994hierarchical}. Context gating introduces gating vectors based on sigmoid activations that are capable of capturing dependencies between input features (input gates) and output labels (output gates). The output then goes through another context gating transformation for the final prediction for each expression.

\begin{figure}
\begin{center}
  \includegraphics[width=1.0\linewidth]{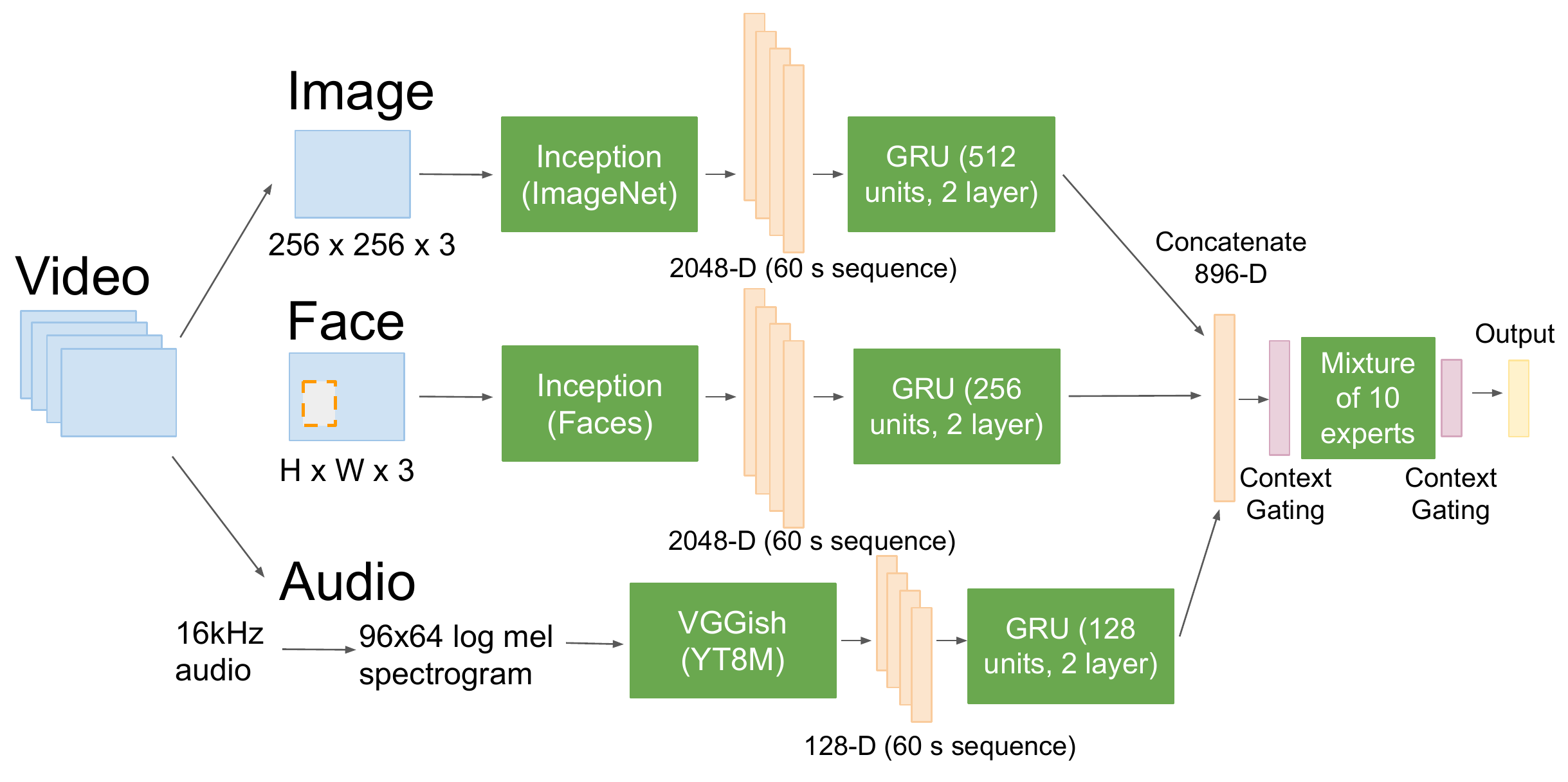}
  \caption{Baseline model architecture with input modalities image, face and audio, based on \cite{sungla}.}
  \label{fig:8}
\end{center}
\end{figure}

\paragraph{Implementation Details.} Our models are trained using the Adam optimizer \cite{kingma2014adam} with a  mini-batch size of 128. We use a learning rate of $0.0005$. We use gradient clipping when training the network to mitigate potential exploding gradient problems. For the GRU, we apply dropout \cite{srivastava2014dropout} of $0.3$ in each layer. For the context gating implementation, we apply batch normalization \cite{ioffe2015batch} before the non-linear layer. The models are implemented in TensorFlow.

\begin{table}
\begin{center}
\begin{tabular}{l c c c c}
\hline
Input & Correlation \\
\hline
Image+Face+Audio &  \textbf{0.139}  \\
Image+Face &  0.114 \\
Face+Audio & 0.126 \\
Image+Audio & \textbf{0.139} \\
Image & 0.118 \\
Face & 0.0893 \\
Audio & 0.123 \\
\hline
\end{tabular}
\end{center}
\caption{Results of the baseline model on EEV with different input modalities. The results are averaged over expressions.}
\label{table:eev_average}
\end{table}

\begin{table*}[t]
\small
\begin{center}
\scalebox{0.9}{
\begin{tabular}{l c c c c c c c c c c c c c c c c c}
\hline
Input & Amus. & Ang. & Awe & Conc. & Conf. & Contem. & Content. &  Disap. & Doubt & Elati. & Inter. & Pain & Sadne. & Surpr. & Trium.  \\
\hline
I+F+A & 0.160 & 0.080 & 0.077 & 0.226 & 0.132 & 0.118 & 0.223 &  0.154 & 0.205 & 0.120 & 0.135 & 0.133 & 0.147 & 0.179 & 0.068  \\
I+F & 0.161 & 0.084 & 0.065 & 0.160 & 0.115 & 0.131 & 0.191 &  0.104 & 0.164 & 0.078 &  0.104 & 0.085 & 0.130 & 0.145 & 0.062 \\
F+A & 0.179 & 0.095 & 0.075 & 0.216 & 0.092 & 0.146 & 0.222 & 0.157 & 0.150 & 0.088 &  0.088 & 0.107 & 0.087 & 0.180 & 0.069 \\
I+A & 0.165 &0.095 & 0.071 & 0.235 & 0.142 & 0.156 & 0.251 & 0.135 & 0.222 & 0.105 &  0.115 & 0.127 & 0.106 & 0.181 & 0.070 \\
I & 0.124 & 0.083 & 0.065 & 0.197 & 0.123 & 0.136 & 0.213 & 0.095 & 0.200 & 0.089 &  0.102 & 0.096 & 0.107 & 0.169 & 0.048 \\
F & 0.140 & 0.061 & 0.034 & 0.156 & 0.044 & 0.095 & 0.173  & 0.130 & 0.101 & 0.076 &  0.060 & 0.062 & 0.112 & 0.099 & 0.046 \\
A & 0.149 & 0.084 & 0.069 & 0.188 & 0.077 & 0.119 & 0.242  & 0.150 & 0.158 & 0.089 &  0.110 & 0.123 & 0.128 & 0.165 & 0.064 \\
\hline
\end{tabular}}
\end{center}
\caption{Results of the per-expression correlation values of the baseline model on EEV. For Inputs, ``I" corresponds to Image input, ``F" corresponds to Face input and ``A" stands for Audio input.}
\label{table:eev_res_correlation}
\end{table*}

\subsection{Datasets}

\paragraph{EEV dataset.} On the EEV dataset, we perform regression of each expression at 1 Hz. This is consistent with the labels in the LIRIS-ACCEDE dataset. We measure performance using Pearson's Correlation Coefficient (correlation). We report the average correlation over expressions.

\paragraph{LIRIS-ACCEDE dataset.} This dataset is chosen because it is the largest dataset that focuses on a similar task to the EEV dataset -- predicting the impact of movies on viewers. The LIRIS-ACCEDE dataset \cite{baveye2015deep} is annotated each second with self-reported valence and arousal values. Each of the two dimensions is a continuous value in $[-1,1]$. We use the same metrics as the dataset competition \cite{dellandrea2018mediaeval} to measure performance: mean squared error (MSE) and Pearson's Correlation Coefficient (correlation). The LIRIS-ACCEDE dataset is divided into three parts during download. We report performance on the first part (14 movies) and the rest (40 movies) for training.

\subsection{Results on EEV}\label{sec:eev_results}
We summarize the results of the EEV dataset in Table~\ref{table:eev_average}. From the summary table, we see that we can learn a positive correlation using all combination of modalities. This result demonstrates that information for predicting affective responses can be learned from the EEV dataset, despite challenges outlined in section~\ref{sec:dataset_char}. Our observations align with the statement from \cite{soleymani2014corpus}, in that affective response is not arbitrary and there is video content that can evoke consistent viewer responses. 

The result for each expression is in Table~\ref{table:eev_res_correlation}. The model is able to predict some expressions, such as amusement, concentration, and surprise, with higher correlation than others, such as anger, and triumph. Although our correlation is low for some expressions, we note that the correlation for each expression is positive, which suggests that information can be learned from the video content to predict each individual expression.

\paragraph{Feature ablation.} We use the same architecture described in Figure~\ref{fig:8} for our ablation study and selectively remove input features to test performance for different feature combinations. From Table~\ref{table:eev_average}, we see that Image+Face+Audio performs similar to Image+Audio (without face features). Additionally, face features alone has the lowest correlation in the single modality tests. These observations suggest that the face modality may be less informative for evoked expressions than image or audio features. Comparing the single modality tests, we see that audio features has the highest correlation. We can further observe that by dropping the audio features from Image+Face+Audio, we experience the biggest decrease in correlation. These results show that adding audio features is helpful if we only have image or face features.

From Table~\ref{table:eev_res_correlation}, we see that removing audio features from Image+Face+Audio lowers the correlation of almost all expressions. In particular, the biggest decreases in performance can be observed for concentration, contentment, and disappointment. In contrast, by removing only image features or only face features from the full set, only the correlation of some expressions are lowered. Since the faces are extracted from the image, it is very likely that there are redundant signals in the remaining face or image feature, so the correlations are not lowered as much. When we remove both image and face features and use audio features alone, the correlations are generally lower for all expressions.

\subsection{Results on LIRIS-ACCEDE}

\begin{table}
\begin{center}
\small
\scalebox{0.9}{
\begin{tabular}{l c c c c}
\hline
Model & V-MSE & A-MSE & V-Correlation & A-Correlation \\
\hline
LIRIS only & 0.091  & 0.104  & 0.216 &  0.223 \\
EEV transfer & \textbf{0.063}  & \textbf{0.071}  & \textbf{0.261}  & \textbf{0.242} \\
\hline
\end{tabular}}
\end{center}
\caption{Results of the baseline model on LIRIS-ACCEDE validation set, using all input features. The mean squared error (MSE) and Correlation metrics preceded with ``V" corresponds to the valence metrics, while the metrics preceded with ``A" corresponds to the arousal metrics. We show results without (first row) and with (second row) transfer learning from EEV.}
\label{table:table3}
\end{table}

The results on the LIRIS-ACCEDE dataset are in Table~\ref{table:table3}. We experiment with training from scratch and fine-tuning from a checkpoint pre-trained on EEV. We use the full feature version of the reaction prediction model with Image+Face+Audio and change the final regression layer to 2 dimensions for valence and arousal. This is to test the ability of EEV to be used for transfer learning in similar tasks.

We observe that the transfer learning model has lower MSE and higher correlation than the model trained from LIRIS only. The improvements to valence metrics are larger than the improvements to arousal. This suggests that pre-training on the EEV data is helpful in improving performance for related tasks.


\section{Conclusion}
We introduce the EEV dataset, a large-scale video dataset for studying evoked viewer facial expressions from video content. EEV is larger and more diverse than previous video datasets for studying viewer reactions. Baseline model performance for predicting evoked expressions shows that while affective information can be learned from video content, there remains a lot of potential for improvement. We hope that the EEV dataset will be useful in developing novel models for affective computing and video analysis. The frame-level evoked expression labels released with EEV can be used to train complex models. Furthermore, we have shown the potential for EEV to be used for transfer learning on related tasks, such as viewer reaction prediction using valence and arousal.

In addition to training new models, the evoked expression labels in EEV can be used to search for themes and videos corresponding to specific expressions. This is useful for finding videos to evoke facial expressions for experiments in psychology and affective computing. Additionally, understanding the distribution of evoked expressions in different themes can help video recommendation systems better cater content to users. Video content creators can also benefit by studying affective stimuli in videos. We hope our evoked expression labels enable further explorations in these directions.


\section{Acknowlegements}

We are grateful to the Computational Vision Lab at Caltech for making this collaboration possible. We would like to thank Marco Andreetto and Brendan Jou from Google
Research for their support and helpful discussions.

\typeout{}
{\small
\bibliographystyle{ieee_fullname}
\bibliography{egbib}

\begin{thebibliography}{10}\itemsep=-1pt

\bibitem{googleio}
Google i/o 2013 - semantic video annotations in the youtube topics api: Theory
  and applications.
\newblock
  \href{https://www.youtube.com/watch?v=wf_77z1H-vQ}{https://www.youtube.com/watch?v=wf\_77z1H-vQ},
  2013.

\bibitem{kg}
Knowledge graph search api.
\newblock
  \href{https://developers.google.com/knowledge-graph/}{https://developers.google.com/knowledge-graph/},
  2015.

\bibitem{yt8m}
Sami Abu{-}El{-}Haija, Nisarg Kothari, Joonseok Lee, Paul Natsev, George
  Toderici, Balakrishnan Varadarajan, and Sudheendra Vijayanarasimhan.
\newblock Youtube-8m: {A} large-scale video classification benchmark.
\newblock {\em CoRR}, abs/1609.08675, 2016.

\bibitem{acar2017comprehensive}
Esra Acar, Frank Hopfgartner, and Sahin Albayrak.
\newblock A comprehensive study on mid-level representation and ensemble
  learning for emotional analysis of video material.
\newblock {\em Multimedia Tools and Applications}, 76(9):11809--11837, 2017.

\bibitem{baveye2018affective}
Yoann Baveye, Christel Chamaret, Emmanuel Dellandr{\'e}a, and Liming Chen.
\newblock Affective video content analysis: A multidisciplinary insight.
\newblock {\em IEEE Transactions on Affective Computing}, 9(4):396--409, 2018.

\bibitem{baveye2015deep}
Yoann Baveye, Emmanuel Dellandr{\'e}a, Christel Chamaret, and Liming Chen.
\newblock Deep learning vs. kernel methods: Performance for emotion prediction
  in videos.
\newblock In {\em 2015 International Conference on Affective Computing and
  Intelligent Interaction (ACII)}, pages 77--83. IEEE, 2015.

\bibitem{baveye2015liris}
Yoann Baveye, Emmanuel Dellandrea, Christel Chamaret, and Liming Chen.
\newblock Liris-accede: A video database for affective content analysis.
\newblock {\em IEEE Transactions on Affective Computing}, 6(1):43--55, 2015.

\bibitem{becker2017emotion}
Hanna Becker, Julien Fleureau, Philippe Guillotel, Fabrice Wendling, Isabelle
  Merlet, and Laurent Albera.
\newblock Emotion recognition based on high-resolution eeg recordings and
  reconstructed brain sources.
\newblock {\em IEEE Transactions on Affective Computing}, 2017.

\bibitem{carreira2017quo}
Joao Carreira and Andrew Zisserman.
\newblock Quo vadis, action recognition? a new model and the kinetics dataset.
\newblock In {\em proceedings of the IEEE Conference on Computer Vision and
  Pattern Recognition}, pages 6299--6308, 2017.

\bibitem{cho2014learning}
Kyunghyun Cho, Bart Van~Merri{\"e}nboer, Caglar Gulcehre, Dzmitry Bahdanau,
  Fethi Bougares, Holger Schwenk, and Yoshua Bengio.
\newblock Learning phrase representations using rnn encoder-decoder for
  statistical machine translation.
\newblock {\em arXiv preprint arXiv:1406.1078}, 2014.

\bibitem{ck2017}
Alan~S. Cowen and Dacher Keltner.
\newblock Self-report captures 27 distinct categories of emotion bridged by
  continuous gradients.
\newblock {\em Proceedings of the National Academy of Sciences},
  114(38):E7900--E7909, 2017.

\bibitem{ck2019}
Alan~S Cowen and Dacher Keltner.
\newblock What the face displays: Mapping 28 emotions conveyed by naturalistic
  expression.
\newblock {\em American Psychologist}, 2019.

\bibitem{dellandrea2018mediaeval}
Emmanuel Dellandr{\'e}a, Liming Chen, Yoann Baveye, Mats~Viktor Sj{\"o}berg,
  Christel Chamaret, et~al.
\newblock The mediaeval 2018 emotional impact of movies task.
\newblock In {\em MediaEval 2018 Multimedia Benchmark Workshop Working Notes
  Proceedings of the MediaEval 2018 Workshop}, 2018.

\bibitem{deng2009imagenet}
Jia Deng, Wei Dong, Richard Socher, Li-Jia Li, Kai Li, and Li Fei-Fei.
\newblock Imagenet: A large-scale hierarchical image database.
\newblock In {\em 2009 IEEE conference on computer vision and pattern
  recognition}, pages 248--255. Ieee, 2009.

\bibitem{deng2017factorized}
Zhiwei Deng, Rajitha Navarathna, Peter Carr, Stephan Mandt, Yisong Yue, Iain
  Matthews, and Greg Mori.
\newblock Factorized variational autoencoders for modeling audience reactions
  to movies.
\newblock In {\em Proceedings of the IEEE Conference on Computer Vision and
  Pattern Recognition}, pages 2577--2586, 2017.

\bibitem{douglas2007humaine}
Ellen Douglas-Cowie, Roddy Cowie, Ian Sneddon, Cate Cox, Orla Lowry, Margaret
  Mcrorie, Jean-Claude Martin, Laurence Devillers, Sarkis Abrilian, Anton
  Batliner, et~al.
\newblock The humaine database: Addressing the collection and annotation of
  naturalistic and induced emotional data.
\newblock In {\em International conference on affective computing and
  intelligent interaction}, pages 488--500. Springer, 2007.

\bibitem{ekman1978manual}
Paul Ekman and Wallace~V Friesen.
\newblock {\em Manual for the facial action coding system}.
\newblock Consulting Psychologists Press, 1978.

\bibitem{gemmeke2017audio}
Jort~F Gemmeke, Daniel~PW Ellis, Dylan Freedman, Aren Jansen, Wade Lawrence,
  R~Channing Moore, Manoj Plakal, and Marvin Ritter.
\newblock Audio set: An ontology and human-labeled dataset for audio events.
\newblock In {\em 2017 IEEE International Conference on Acoustics, Speech and
  Signal Processing (ICASSP)}, pages 776--780. IEEE, 2017.

\bibitem{hanjalic2005affective}
Alan Hanjalic and Li-Qun Xu.
\newblock Affective video content representation and modeling.
\newblock {\em IEEE transactions on multimedia}, 7(1):143--154, 2005.

\bibitem{horvat2013multimedia}
Marko Horvat, Sini{\v{s}}a Popovi{\'c}, and K Cosi{\'c}.
\newblock Multimedia stimuli databases usage patterns: a survey report.
\newblock In {\em 2013 36th International Convention on Information and
  Communication Technology, Electronics and Microelectronics (MIPRO)}, pages
  993--997. IEEE, 2013.

\bibitem{ioffe2015batch}
Sergey Ioffe and Christian Szegedy.
\newblock Batch normalization: Accelerating deep network training by reducing
  internal covariate shift.
\newblock {\em arXiv preprint arXiv:1502.03167}, 2015.

\bibitem{jiang2014predicting}
Yu-Gang Jiang, Baohan Xu, and Xiangyang Xue.
\newblock Predicting emotions in user-generated videos.
\newblock In {\em Twenty-Eighth AAAI Conference on Artificial Intelligence},
  2014.

\bibitem{jordan1994hierarchical}
Michael~I Jordan and Robert~A Jacobs.
\newblock Hierarchical mixtures of experts and the em algorithm.
\newblock {\em Neural computation}, 6(2):181--214, 1994.

\bibitem{kahou2013combining}
Samira~Ebrahimi Kahou, Christopher Pal, Xavier Bouthillier, Pierre Froumenty,
  {\c{C}}aglar G{\"u}l{\c{c}}ehre, Roland Memisevic, Pascal Vincent, Aaron
  Courville, Yoshua Bengio, Raul~Chandias Ferrari, et~al.
\newblock Combining modality specific deep neural networks for emotion
  recognition in video.
\newblock In {\em Proceedings of the 15th ACM on International conference on
  multimodal interaction}, pages 543--550. ACM, 2013.

\bibitem{karpathy2014large}
Andrej Karpathy, George Toderici, Sanketh Shetty, Thomas Leung, Rahul
  Sukthankar, and Li Fei-Fei.
\newblock Large-scale video classification with convolutional neural networks.
\newblock In {\em Proceedings of the IEEE conference on Computer Vision and
  Pattern Recognition}, pages 1725--1732, 2014.

\bibitem{KassamKarim2011Aoee}
Karim Kassam.
\newblock Assessment of emotional experience through facial expression, 2011.

\bibitem{kingma2014adam}
Diederik~P Kingma and Jimmy Ba.
\newblock Adam: A method for stochastic optimization.
\newblock {\em arXiv preprint arXiv:1412.6980}, 2014.

\bibitem{koelstra2012deap}
Sander Koelstra, Christian Muhl, Mohammad Soleymani, Jong-Seok Lee, Ashkan
  Yazdani, Touradj Ebrahimi, Thierry Pun, Anton Nijholt, and Ioannis Patras.
\newblock Deap: A database for emotion analysis; using physiological signals.
\newblock {\em IEEE transactions on affective computing}, 3(1):18--31, 2012.

\bibitem{lowe2004distinctive}
David~G Lowe.
\newblock Distinctive image features from scale-invariant keypoints.
\newblock {\em International journal of computer vision}, 60(2):91--110, 2004.

\bibitem{thugla}
Ye Ma, Xihao Liang, and Mingxing Xu.
\newblock Thuhcsi in mediaeval 2018 emotional impact of movies task.
\newblock 2018.

\bibitem{martinez2017automatic}
Brais Martinez, Michel~F Valstar, Bihan Jiang, and Maja Pantic.
\newblock Automatic analysis of facial actions: A survey.
\newblock {\em IEEE Transactions on Affective Computing}, 2017.

\bibitem{mcduff2015predicting}
Daniel McDuff, Rana El~Kaliouby, Jeffrey~F Cohn, and Rosalind~W Picard.
\newblock Predicting ad liking and purchase intent: Large-scale analysis of
  facial responses to ads.
\newblock {\em IEEE Transactions on Affective Computing}, 6(3):223--235, 2015.

\bibitem{mcduff2013predicting}
Daniel McDuff, Rana El~Kaliouby, David Demirdjian, and Rosalind Picard.
\newblock Predicting online media effectiveness based on smile responses
  gathered over the internet.
\newblock In {\em 2013 10th IEEE international conference and workshops on
  automatic face and gesture recognition (FG)}, pages 1--7. IEEE, 2013.

\bibitem{mcduff2013affectiva}
Daniel McDuff, Rana Kaliouby, Thibaud Senechal, May Amr, Jeffrey Cohn, and
  Rosalind Picard.
\newblock Affectiva-mit facial expression dataset (am-fed): Naturalistic and
  spontaneous facial expressions collected.
\newblock In {\em Proceedings of the IEEE Conference on Computer Vision and
  Pattern Recognition Workshops}, pages 881--888, 2013.

\bibitem{mcduff2017large}
Daniel McDuff and Mohammad Soleymani.
\newblock Large-scale affective content analysis: Combining media content
  features and facial reactions.
\newblock In {\em 2017 12th IEEE International Conference on Automatic Face \&
  Gesture Recognition (FG 2017)}, pages 339--345. IEEE, 2017.

\bibitem{miech2017learnable}
Antoine Miech, Ivan Laptev, and Josef Sivic.
\newblock Learnable pooling with context gating for video classification.
\newblock {\em arXiv preprint arXiv:1706.06905}, 2017.

\bibitem{monfort2019moments}
Mathew Monfort, Alex Andonian, Bolei Zhou, Kandan Ramakrishnan, Sarah~Adel
  Bargal, Yan Yan, Lisa Brown, Quanfu Fan, Dan Gutfreund, Carl Vondrick, et~al.
\newblock Moments in time dataset: one million videos for event understanding.
\newblock {\em IEEE transactions on pattern analysis and machine intelligence},
  2019.

\bibitem{poria2015towards}
Soujanya Poria, Erik Cambria, Amir Hussain, and Guang-Bin Huang.
\newblock Towards an intelligent framework for multimodal affective data
  analysis.
\newblock {\em Neural Networks}, 63:104--116, 2015.

\bibitem{schaefer2010assessing}
Alexandre Schaefer, Fr{\'e}d{\'e}ric Nils, Xavier Sanchez, and Pierre
  Philippot.
\newblock Assessing the effectiveness of a large database of emotion-eliciting
  films: A new tool for emotion researchers.
\newblock {\em Cognition and Emotion}, 24(7):1153--1172, 2010.

\bibitem{Schroff_2015_CVPR}
Florian Schroff, Dmitry Kalenichenko, and James Philbin.
\newblock Facenet: A unified embedding for face recognition and clustering.
\newblock In {\em The IEEE Conference on Computer Vision and Pattern
  Recognition (CVPR)}, June 2015.

\bibitem{soleymani2016analysis}
Mohammad Soleymani, Sadjad Asghari-Esfeden, Yun Fu, and Maja Pantic.
\newblock Analysis of eeg signals and facial expressions for continuous emotion
  detection.
\newblock {\em IEEE Transactions on Affective Computing}, 7(1):17--28, 2016.

\bibitem{soleymani2009bayesian}
Mohammad Soleymani, Joep~JM Kierkels, Guillaume Chanel, and Thierry Pun.
\newblock A bayesian framework for video affective representation.
\newblock In {\em 2009 3rd International Conference on Affective Computing and
  Intelligent Interaction and Workshops}, pages 1--7. IEEE, 2009.

\bibitem{soleymani2014corpus}
Mohammad Soleymani, Martha Larson, Thierry Pun, and Alan Hanjalic.
\newblock Corpus development for affective video indexing.
\newblock {\em IEEE Transactions on Multimedia}, 16(4):1075--1089, 2014.

\bibitem{soleymani2012multimodal}
Mohammad Soleymani, Jeroen Lichtenauer, Thierry Pun, and Maja Pantic.
\newblock A multimodal database for affect recognition and implicit tagging.
\newblock {\em IEEE Transactions on Affective Computing}, 3(1):42--55, 2012.

\bibitem{srivastava2014dropout}
Nitish Srivastava, Geoffrey Hinton, Alex Krizhevsky, Ilya Sutskever, and Ruslan
  Salakhutdinov.
\newblock Dropout: a simple way to prevent neural networks from overfitting.
\newblock {\em The Journal of Machine Learning Research}, 15(1):1929--1958,
  2014.

\bibitem{sungla}
Jennifer~J Sun, Ting Liu, and Gautam Prasad.
\newblock Gla in mediaeval 2018 emotional impact of movies task.
\newblock {\em arXiv preprint arXiv:1911.12361}, 2019.

\bibitem{szegedy2016rethinking}
Christian Szegedy, Vincent Vanhoucke, Sergey Ioffe, Jon Shlens, and Zbigniew
  Wojna.
\newblock Rethinking the inception architecture for computer vision.
\newblock In {\em Proceedings of the IEEE conference on computer vision and
  pattern recognition}, pages 2818--2826, 2016.

\bibitem{Vemulapalli_2019_CVPR}
Raviteja Vemulapalli and Aseem Agarwala.
\newblock A compact embedding for facial expression similarity.
\newblock In {\em The IEEE Conference on Computer Vision and Pattern
  Recognition (CVPR)}, June 2019.

\bibitem{wang2006affective}
Hee~Lin Wang and Loong-Fah Cheong.
\newblock Affective understanding in film.
\newblock {\em IEEE Transactions on circuits and systems for video technology},
  16(6):689--704, 2006.

\bibitem{wang2015video}
Shangfei Wang and Qiang Ji.
\newblock Video affective content analysis: a survey of state-of-the-art
  methods.
\newblock {\em IEEE Transactions on Affective Computing}, 6(4):410--430, 2015.

\bibitem{zhang2009utilizing}
Shiliang Zhang, Qi Tian, Qingming Huang, Wen Gao, and Shipeng Li.
\newblock Utilizing affective analysis for efficient movie browsing.
\newblock In {\em 2009 16th IEEE International Conference on Image Processing
  (ICIP)}, pages 1853--1856. IEEE, 2009.

\bibitem{zhao2013video}
Sicheng Zhao, Hongxun Yao, and Xiaoshuai Sun.
\newblock Video classification and recommendation based on affective analysis
  of viewers.
\newblock {\em Neurocomputing}, 119:101--110, 2013.

\bibitem{zlatintsi2017cognimuse}
Athanasia Zlatintsi, Petros Koutras, Georgios Evangelopoulos, Nikolaos
  Malandrakis, Niki Efthymiou, Katerina Pastra, Alexandros Potamianos, and
  Petros Maragos.
\newblock Cognimuse: A multimodal video database annotated with saliency,
  events, semantics and emotion with application to summarization.
\newblock {\em EURASIP Journal on Image and Video Processing}, 2017(1):54,
  2017.

\end{thebibliography}
}

\end{document}